\documentclass[letterpaper]{article} 
\usepackage{aaai25}  
\usepackage{times}  
\usepackage{helvet}  
\usepackage{courier}  
\usepackage{booktabs} 
\usepackage{multirow} 
\usepackage{graphicx}
\usepackage{amssymb} 
\usepackage{amsthm}  
\theoremstyle{definition}
\newtheorem{definition}{Definition}
\usepackage{svg} 
\usepackage{amsmath} 
\usepackage[hyphens]{url}  
\usepackage{graphicx} 
\usepackage{natbib}  
\usepackage{caption} 
\usepackage{algorithm}
\usepackage{algorithmic}
\usepackage{amsmath} 
\usepackage{amsthm} 

\usepackage{newfloat}
\usepackage{listings}
\DeclareCaptionStyle{ruled}{labelfont=normalfont,labelsep=colon,strut=off} 
\floatstyle{ruled}
\newfloat{listing}{tb}{lst}{}
\floatname{listing}{Listing}
\title{Towards Robust Uncertainty-Aware Incomplete Multi-View Classification}
\author{
    Mulin Chen\textsuperscript{\rm 1}\thanks{Corresponding author}, 
    Haojian Huang\textsuperscript{\rm 2}\footnotemark[1], 
    Qiang Li\textsuperscript{\rm 1} 
}
\affiliations{
    \textsuperscript{\rm 1}School of Computer Science, Northwestern Polytechnical University, Xi'an, China\\
    \textsuperscript{\rm 2}Department of Computer Science, The University of Hong Kong, Hong Kong, China\\
    chenmulin001@gmail.com, haojianhuang@connect.hku.hk, mozhu87@mail.nwpu.edu.cn
}

\usepackage{bibentry}
\usepackage{lipsum}

\begin{document}

\maketitle

\begin{abstract}
Handling incomplete data in multi-view classification is challenging, especially when traditional imputation methods introduce biases that compromise uncertainty estimation. Existing Evidential Deep Learning (EDL) based approaches attempt to address these issues, but they often struggle with conflicting evidence due to the limitations of the Dempster-Shafer combination rule, leading to unreliable decisions. To address these challenges, we propose the Alternating Progressive Learning Network (APLN), specifically designed to enhance EDL-based methods in incomplete MVC scenarios. Our approach mitigates bias from corrupted observed data by first applying coarse imputation, followed by mapping the data to a latent space. In this latent space, we progressively learn an evidence distribution aligned with the target domain, incorporating uncertainty considerations through EDL. Additionally, we introduce a conflict-aware Dempster-Shafer combination rule (DSCR) to better handle conflicting evidence. By sampling from the learned distribution, we optimize the latent representations of missing views, reducing bias and enhancing decision-making robustness. Extensive experiments demonstrate that APLN, combined with DSCR, significantly outperforms traditional methods, particularly in environments characterized by high uncertainty and conflicting evidence, establishing it as a promising solution for incomplete multi-view classification.
\end{abstract}

%

\section{Introduction}
\label{sec:intro}
In the field of multi-view classification, handling incomplete data has always been a significant challenge. For the task of incomplete multi-view classification (IMVC), existing studies can be broadly divided into two primary categories. The first category comprises methods that perform classification using only the available views without imputing the missing data. Although these methods~\cite{DeepIMV, CPM} avoid the complexity of data reconstruction, they often struggle when faced with high missing rates, as they fail to fully exploit the correlations between views. Consequently, these approaches are typically less effective and robust in scenarios where a significant portion of the data is missing. The second category of methods~\cite{MVAE, MIWAE, impute_ADMM, impute_adversarial} seeks to reconstruct the missing data using deep learning techniques, such as autoencoders~\cite{AE1, AE2} or generative adversarial networks (GANs)~\cite{GAN}, before performing classification on the imputed complete data. While these approaches have shown considerable promise, they are not without drawbacks. Primarily, the deterministic nature of these methods fails to adequately capture the uncertainty associated with the missing data, leading to potentially unstable classification outcomes. To mitigate this issue, multi-value imputation, as opposed to single-value imputation, has demonstrated an ability to produce uncertainty-aware predictions, resulting in more stable downstream performance. These approaches typically sample several reference samples from the complete data distribution in the observed space as imputation candidates for the missing samples. However, the imputation process often lacks interpretability, making it difficult to understand the underlying mechanisms. 
\begin{figure}
    \centering
    \includegraphics[width=\linewidth]{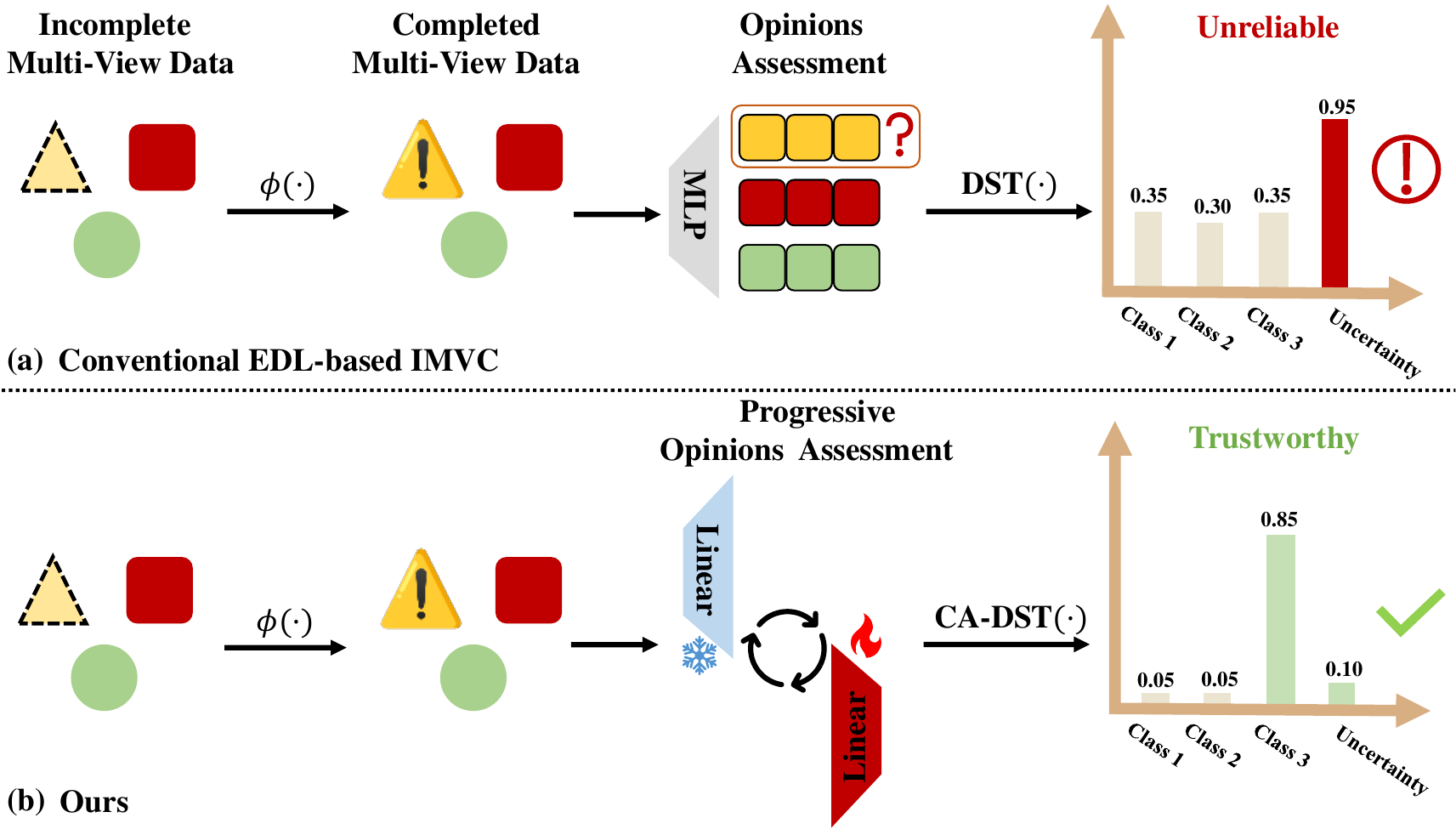}
    \vspace{-2mm}
    \caption{Comparison between traditional EDL-based methods and our approach for handling IMVC. While traditional methods often result in high uncertainty due to inter-view conflicts after completing the missing views, our method leverages conflict-aware Dempster-Shafer (CA-DST) for progressive refinement, leading to more reliable inference and significantly reduced uncertainty.} 
    \label{fig:comp}
    \vspace{-6mm}
\end{figure}
Existing methods seek to address this challenge by utilizing quality assessment models that evaluate the bias and uncertainty introduced by imputation within each view's representation, thereby enhancing downstream decision-making. A widely recognized approach for uncertainty estimation is Evidential Deep Learning (EDL), whose effectiveness has been demonstrated in methods like UIMC~\cite{xie2023exploring}. 
Nevertheless, due to the inherent bias introduced by missing data, the distribution in the observed space does not accurately represent the true target distribution. Consequently, the imputation references sampled from this biased distribution can be biased, leading to suboptimal performance. What is worse, suboptimal imputation outcomes can distort view representations, potentially introducing noise or even conflicts. The vallina EDL-based quality assessment models are actually constrained by the Dempster-Shafer combination rule (DSCR), which is highly sensitive to conflicting evidence, often resulting in counterintuitive and anomalous outcomes that lead to suboptimal decisions. When conflicts arise in the imputed results, the limitations of the DSCR hinder further collaborative optimization of the imputed view representations. The absence of latent space refinement for imputed views limits the generation of corrective feedback needed to resolve conflicts during representation learning and decision-making.

To tackle the challenges of bias and interpolation uncertainty in corrupted observed domains, we propose an Alternating Progressive Learning Network (APLN). This method systematically addresses distributional bias by progressively refining the observed space and aligning it with the target domain. Initially, we perform a coarse imputation of the observed domain, followed by training a variational autoencoder (VAE) to learn a latent space that partially aligns with the target domain. In the second phase, we employ EDL to model a Dirichlet distribution sensitive to uncertainties from the observed space, bringing the learned distribution closer to the target domain. To prevent degradation due to potential conflicts, we introduce a novel consistency loss using a conflict-aware DSCR, enhancing the robustness of the evidence-based Dirichlet distribution. Finally, in the third phase, we jointly refine the latent space and evidence distributions, achieving a comprehensive alignment with the target domain. This approach not only improves the latent representation of incomplete views but also ensures more reliable and robust decision-making. Our contributions are:
\begin{itemize}
    \item Our proposed method adopts a progressive learning strategy that iteratively aligns the latent representations from the observed and feature spaces with the target domain. By incorporating uncertainty-sensitive Dirichlet modeling in the second phase, our approach effectively captures complex data distribution shifts between the observed and target domains, thereby enhancing the accuracy and trustworthiness of decision-making.
    \item We introduce a novel consistency loss based on a conflict-aware DSCR to enhance the robustness of the evidence-based Dirichlet distribution in the presence of conflicting observations.
    \item We validate the effectiveness of our proposed method through extensive experiments, demonstrating significant improvements over traditional methods, particularly in scenarios with high uncertainty and conflicts.
\end{itemize}

\section{Related Work}
\subsection{Incomplete Multi-View Learning}
Incomplete multi-view learning, a central challenge in multi-view classification, focuses on effectively managing missing views. Existing methods fall into two categories: 1) approaches that work directly on available views, learning a common latent representation without reconstructing missing data~\cite{MCP, PMC, IMVSG, CPM, DeepIMV}; and 2) generative methods that impute missing views before downstream tasks, using techniques like VAEs and GANs~\cite{MVAE, MVTCAE, MIWAE, CPM-GAN, GAN_clu1, GAN_clu2}. However, both approaches have limitations. The former often suffers from performance degradation due to limited view correlations, while the latter tends to rely on deterministic imputation~\cite{MCWIV, IMVC, COMPLETER}, failing to capture the inherent uncertainty of missing data, which can result in unstable outcomes, particularly in high-dimensional scenarios. To address these challenges, we propose a multi-stage interpolation approach that refines imputation by first performing coarse interpolation in the observed space, followed by further refinement in the latent space. This allows for better uncertainty assessment, enabling more robust inference.

Moreover, while EDL~\cite{sensoy2018evidential} has been effective in modeling uncertainty, its application to incomplete multi-view classification (IMVC) is hindered by sensitivity to evidence conflicts in the observed domain. To mitigate this, we introduce a conflict-aware DSCR with a new consistency loss, ensuring more coherent evidence fusion and improved performance in IMVC tasks.
\subsection{Uncertianty-Based Deep Learning}
 Numerous studies have focused on developing models capable of estimating uncertainty to enhance reliability and trustworthiness in decision-making~\cite{zhang2021bayesian, xiao2021bit, pinoise, izmailov2021bayesian, gal2016dropout, amini2020deep, sensoy2018evidential,liu2023adaptive,liu2023safe,chen2024finecliper,ma2024beyond}. Among these, EDL~\cite{sensoy2018evidential} has gained attention for its ability to model "second-order probabilities" over logits using Dempster-Shafer Theory~\cite{shafer1992dempster} and Subjective Logic~\cite{jsang2018subjective}. This approach captures uncertainty conveniently and accurately across various domains~\cite{qin2022deep, shao2024dual, holmquist2023evidential, huang2024trusted, huang2024crest, huang2024evidential}. However, when applying EDL-based methods to IMVC (\emph{e.g.}~\cite{xie2023exploring}), existing approaches often overlook the uncertainty introduced by the corrupted nature of the observed domain. Relying on samples drawn from the observed distribution for imputation can inadvertently introduce conflicts. This is a notable weakness of vanilla EDL, as the DSCR is highly sensitive to evidence conflicts~\cite{huang2023belief}. Even a single piece of conflicting evidence can lead to anomalous fusion results, ultimately degrading the performance of downstream IMVC tasks. To this end, we propose a novel conflict-aware DSCR, incorporating a new consistency loss to achieve more coherent evidence fusion.
\begin{figure*}
    \centering
    \includegraphics[width=\linewidth]{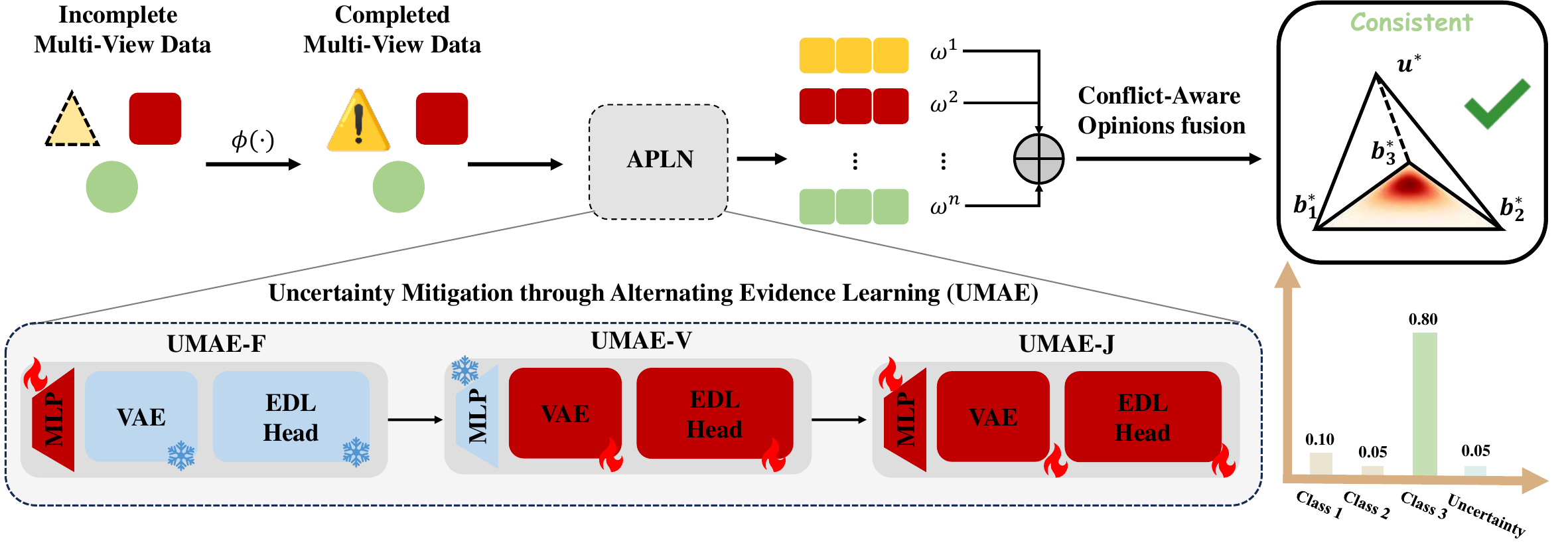}
    \vspace{-4mm}
    \caption{Workflow of ALPN.} 
    \label{fig:comp}
    \vspace{-4mm}
\end{figure*}
\section{Method}
Our method focuses on leveraging correlations among multiple views while addressing uncertainty and conflicts due to missing data, ultimately improving multi-view fusion classification. We first define IMVC and introduce the Uncertainty Mitigation through Alternating Evidence Learning (UMAE) model architecture, then discuss the training details, and finally, explore the reasons behind the model's effectiveness.

\subsection{Background}
Given $N$ training inputs $\left\{\mathbb{X}_n\right\}_{n=1}^N$ with $V$ views, i.e., $\mathbb{X}=\left\{\mathbf{x}^v\right\}_{v=1}^V$, and the corresponding class labels $\left\{\mathbf{y}_n\right\}_{n=1}^N$, multi-view classification aims to construct a mapping between input and label by exploiting the complementary multi-view data. In this paper, we focus on the IMVC task defined in Definition~\ref{def:IMVC}


\setcounter{definition}{0}
\begin{definition}
    \label{def:IMVC}
    \textbf{Incomplete Multi-View Classification.}
    A complete multi-view sample is composed of $V$ views $\mathbb{X} = \left\{\mathbf{x}^v\right\}_{v=1}^V$ and the corresponding class label $\mathbf{y}$. An incomplete multi-view observation $\overline{\mathbb{X}}$ is a subset of the complete multi-view observation (i.e., $\overline{\mathbb{X}} \subseteq \mathbb{X}$) with arbitrary possible $\bar{V}$ views, where $1 \leq \bar{V} \leq V$. Given an incomplete multi-view training dataset $\left\{\overline{\mathbb{X}}_n, \mathbf{y}_n\right\}_{n=1}^N$ with $N$ samples, IMVC aims to learn a mapping between the incomplete multi-view observation $\overline{\mathbb{X}}$ and the corresponding class label $\mathbf{y}$.
\end{definition}

\subsection{UMAE Model Architecture}
\subsubsection{Imputation of Missing Views.}
We first perform random imputation on the missing view data, then project each view \(\mathbf{x}^v\) into a unified feature space using a set of linear transformations \(\{f_c^v\}_{v=1}^V: \{\mathbf{x}^v\} \mapsto \{\mathbf{z}^v\}\). Within this feature space, random masks \(\mathbf{m}_n=\left\{m_n^v\right\}_{v=1}^V\) are generated to simulate missing features, with the missing parts of \(\mathbf{z}_n\) set to 0 to obtain the masked features. By concatenating the masked features with the mask \(\mathbf{m}_n\), the input \(\tilde{\mathbf{z}}_n\) for the VAE model is obtained.

The VAE model takes \(\tilde{\mathbf{z}}_n\) as input and outputs \(\hat{\mathbf{z}}\), with the ground truth features denoted as \(\mathbf{z}\). Training the VAE involves minimizing the negative Evidence Lower Bound (ELBO), represented by the loss function \(\mathcal{L}_{\text{ELBO}}\). The ELBO consists of two key components: a reconstruction error term \(\mathbb{E}_{q(\hat{\mathbf{z}}|\tilde{\mathbf{z}}_n)}[\log p(\mathbf{z}|\hat{\mathbf{z}})]\), which measures the log-likelihood of the true features \(\mathbf{z}\) given the model's latent variables \(\hat{\mathbf{z}}\), and a KL divergence term \(\text{KL}(q(\hat{\mathbf{z}}|\tilde{\mathbf{z}}_n) \parallel p(\mathbf{z}))\), which quantifies the divergence between the approximate posterior \(q(\hat{\mathbf{z}}|\tilde{\mathbf{z}}_n)\) and the prior \(p(\mathbf{z})\). The ELBO is thus defined as:
\begin{equation}
\label{eq:ELBO}
\text{ELBO} = \mathbb{E}_{q(\hat{\mathbf{z}}|\tilde{\mathbf{z}}_n)}\left[\log p(\mathbf{z}|\hat{\mathbf{z}})\right] - \text{KL}\left(q(\hat{\mathbf{z}}|\tilde{\mathbf{z}}_n) \parallel p(\mathbf{z})\right),
\end{equation}
where \(p(\mathbf{z})\) represents the prior distribution, often a standard normal distribution \(N(0, I)\), while \(q(\hat{\mathbf{z}}|\tilde{\mathbf{z}}_n)\) denotes the approximate posterior distribution parameterized by a neural network. \(p(\mathbf{z}|\hat{\mathbf{z}})\) is the distribution used to reconstruct the observed data from the latent space.

To enhance robustness, we sample multiple imputations from the learned distribution, offering a comprehensive approach to handling uncertainty in missing views. By projecting views into a unified feature space, the method effectively learns a reliable model that leverages feature correlations.

\subsubsection{Multi-View Opinion Fusion.}
Formally, let the VAE multiple sampling outputs for view $v$ be denoted as $\hat{\mathbf{z}}^v$, where $\hat{\mathbf{z}}^v$ represents the reconstructed feature for the $v$-th view. We then create the reconstructed feature $\mathbf{z}_{\mathrm{rc}}^n$ for each sample $n$ by combining the observed and reconstructed features. For each sample $n$, the reconstructed feature is computed as:
\begin{equation}
\mathbf{z}_{\mathrm{rc}}^v=m^v \cdot \mathbf{z}^v+\left(1-m^v\right) \cdot \hat{\mathbf{z}}^v
\end{equation}

These features are then projected into the evidence space using a linear transformation $\{f_e^v\}_{v=1}^V$

\begin{equation}
\mathbf{e}^v=f_e^v\left(\mathbf{z}_{\mathrm{rc}}^v\right)
\end{equation}
where $f_v^e$ is a linear layer that maps the reconstructed features into the evidence space.

For $K$ classification problems, the multinomial opinion over a specific view of an instance $\left(\mathbf{x}_n^v\right)^2$ is represented as a triplet $w=(\boldsymbol{b}, u, \boldsymbol{a})$. Here, $\boldsymbol{b}=\left(b_1, \ldots, b_k\right)^{\top}$ denotes the belief masses assigned to possible values based on evidence support, $u$ represents the uncertainty mass reflecting evidence ambiguity, and $\boldsymbol{a}=\left(a_1, \ldots, a_k\right)^{\top}$ is the prior probability distribution for each class. According to subjective logic, both $\boldsymbol{b}$ and $u$ must be non-negative and their sum must equal one:

\begin{equation}
\sum_{k=1}^K b_k+u=1, \forall k \in[1, \ldots, K]
\end{equation}

where $b_k \geq 0$ and $u \geq 0$. The projected probability distribution of multinomial opinions is given by:

\begin{equation}
P_k=b_k+a_k u, \forall k \in[1, \ldots, K]
\end{equation}

On opinion aggregation, let \( w^A = (b_k^A, u^A, a^A) \) and \( w^B = (b_k^B, u^B, a^B) \) be the opinions of views \( A \) and \( B \) over the same instance, respectively. The conflictive aggregated opinion \( w^{A \oplus B} \) is calculated as follows:

\begin{equation}
\boldsymbol{\omega}^{A\oplus B} = \boldsymbol{\omega}^A \oplus \boldsymbol{\omega}^B = \left(b_k^{A\oplus B}, u^{A\oplus B}, a^{A\oplus B}\right),
\end{equation}

\begin{equation}
b_k^{A\oplus B} = \frac{b_k^A u^B + b_k^B u^A}{u^A + u^B},
\end{equation}

\begin{equation}
u^{A\oplus B} = \frac{2u^A u^B}{u^A + u^B}, \quad a_k^{A\oplus B} = \frac{a_k^A}{2} + \frac{a_k^B}{2}.
\end{equation}

The opinion \(\boldsymbol{\omega}^{A\oplus B}\) is equivalent to averaging the view-specific evidences:

\begin{equation}
e^{A \oplus B} = \frac{1}{2} \left( e^A + e^B \right).
\end{equation}

represents the combination of the dependent opinions of \(A\) and \(B\). This combination is achieved by mapping the belief opinions to evidence opinions using a bijective mapping between multinomial opinions and the Dirichlet distribution. Essentially, the combination rule ensures that the quality of the new opinion is proportional to the combined one. In other words, when a highly uncertain opinion is combined, the uncertainty of the new opinion is larger than the original opinion. 

Then,we can fuse the final joint opinions \(w\) from different views with the following rule:

\begin{equation}
\boldsymbol{\omega} = \boldsymbol{\omega}^1 \oplus \boldsymbol{\omega}^2 \oplus \dots \oplus \boldsymbol{\omega}^V.
\end{equation}

According to the above fusion rules, we can get the finalmulti-view joint opinion, and thus get the final probabilityof each class and the overall uncertainty.

\subsubsection{Conflict-Aware Consistency Loss.}
We introduce a new conflict degree measure that unifies probability distribution and uncertainty, ensuring a symmetric measure within the range [0, 1]. The new distributions are defined as \( q^A_k = p^A_k (1 - u^A) \) and \( q^B_k = p^B_k (1 - u^B) \). Using the Jensen-Shannon divergence:
\begin{equation}
D_{\text{JS}}(q^A \| q^B) = \frac{1}{2} D_{\text{KL}}(q^A \| M) + \frac{1}{2} D_{\text{KL}}(q^B \| M)
\end{equation}
where \( M = \frac{1}{2}(q^A + q^B) \). The KL divergence is given by \( D_{\text{KL}}(q^A \| M) = \sum_{k=1}^{K} q^A_k \log \frac{q^A_k}{M_k} \) and \( D_{\text{KL}}(q^B \| M) = \sum_{k=1}^{K} q^B_k \log \frac{q^B_k}{M_k} \). The new conflict degree measure is then:

\begin{equation}
c(\boldsymbol{\omega}^A, \boldsymbol{\omega}^B) = 1 - D_{\text{JS}}(q^A \| q^B)
\end{equation}

Thus, the conflict loss for multi-view fusion is:
\begin{equation}
\mathcal{L}_{\text{con}} = \frac{1}{V-1} \sum_{A=1}^{V} \left( \sum_{B \neq A}^{V} c\left(\boldsymbol{\omega}^{A}, \boldsymbol{\omega}^{B}\right) \right).
\end{equation}

The proposed conflictive degree measure improves upon the original by ensuring symmetry through the Jensen-Shannon divergence, which keeps the conflict assessment balanced regardless of input order. It operates within a clear range of [0, 1], with 0 representing maximum conflict and 1 indicating no conflict, thus providing an intuitive and interpretable scale. By unifying probability distributions and uncertainties into a single metric, the measure offers a more comprehensive understanding of conflict, directly reflecting the degree of disagreement between opinions in a way that is both holistic and easy to interpret.

\subsection{Alternating Pro-Gressive Learning Network}
To optimize the performance of our model, we developed an Alternating Progressive Learning Network, which consists of three phases: Feature Training Phase (UMAE-F), VAE-EDL Training Phase (UMAE-V), Joint Training Phase (UMAE-J). The model parameters \(\theta\) consist of the following components: \(\theta_c\) for the feature extraction module, \(\theta_e\) for the EDL module,and \(\theta_v\) for the VAE module.

\subsubsection{UMAE-F.}
In this phase, we use noise to fill in the missing feature views and leverage labels for coarse alignment of the feature space, optimizing \(\theta_c\). This step accelerates the subsequent alignment between latent and target distribution. The loss function \(\mathcal{L}_{\text{acc}}\) is used and we firstly introduced \(\mathcal{L}_{\text{ace}}\) as below:
\begin{equation}
\begin{split}
    \mathcal{L}_{\text{ace}}(\alpha_n) &= \int \left[ \sum_{j=1}^{K} -y_{nj} \log p_{nj} \right] \frac{\sum_{k=1}^{K} \alpha_{nj} - 1}{B(\alpha_n)} \, dp_n \\
    &= \sum_{j=1}^{K} y_{nj} \left( \psi(S_n) - \psi(\alpha_{nj}) \right)
\end{split}
\end{equation}
where \(\psi(\cdot)\) is the digamma function. Nevertheless, this loss function does not ensure lower evidence for incorrect labels. To address this, we introduce the Kullback-Leibler (KL) divergence:
\begin{equation}
\begin{split}
\mathcal{L}_{\text{KL}}(\alpha_n) &= \text{KL}\left[ D(p_n \| \hat{\alpha}_n) \| D(p_n \| 1) \right] \\
&= \log \left[ \frac{\Gamma\left(\sum_{k=1}^{K} \hat{\alpha}_{nk}\right)}{\Gamma(K)\prod_{k=1}^{K} \Gamma(\hat{\alpha}_{nk})} \right] \\
&\quad + \sum_{k=1}^{K} \left( \hat{\alpha}_{nk} - 1 \right) 
\left[ \psi(\hat{\alpha}_{nk}) - \psi\left( \sum_{j=1}^{K} \hat{\alpha}_{nj} \right) \right],
\end{split}
\end{equation}
where \(D(p_n \| 1)\) is the uniform Dirichlet distribution, \(\hat{\alpha}_n = y_n + (1 - y_n)\odot\alpha_n\) is the Dirichlet parameter after removal of the non-misleading evidence from predicted parameters \(\alpha_n\) for the \(n\)-th instance, and \(\Gamma(\cdot)\) is the gamma function.

Therefore, given the Dirichlet distribution with parameter \(\alpha_n\) for the \(n\)-th instance, the loss function is defined as:
\begin{equation}
\mathcal{L}_{\text{acc}}(\alpha_n) = \mathcal{L}_{\text{ace}}(\alpha_n) + \lambda_t \mathcal{L}_{\text{KL}}(\alpha_n),
\end{equation}

\subsubsection{UMAE-V.}
In this stage, we freeze the pre-trained \(\{f_v\}_{v=1}^V\) and use EDL to reduce the bias in the latent space representations introduced by the observed domain. The loss function used here optimizes \(\theta_v\) and \(\theta_e\):

\begin{equation}
\mathcal{L}_{\text{edl}} = \mathcal{L}_{\text{acc}} + \mathcal{L}_{\text{con}},
\end{equation}

\subsubsection{UMAE-J.}
In UMAE-J, we unfreeze the feature layers and introduce a VAE reconstruction loss, enabling all model components to coordinate effectively for optimal performance. The combined loss optimizes \(\theta_c\), \(\theta_v\), and \(\theta_e\):

\begin{equation}
\mathcal{L}_{\text{J}} = \mathcal{L}_{\text{edl}} + \mathcal{L}_{\text{ELBO}},
\end{equation}

where
\begin{equation}
\begin{split}
    \mathcal{L}_{\text{ELBO}} &= - \mathbb{E}_{q(\hat{\mathbf{z}}|\tilde{\mathbf{z}}_n)}\left[\log p(\mathbf{z}|\hat{\mathbf{z}})\right] \\
    &\quad + \text{KL}\left(q(\hat{\mathbf{z}}|\tilde{\mathbf{z}}_n) \parallel p(\mathbf{z})\right)
\end{split}
\end{equation}

\section{Experiment}
\label{Experiments}
\subsection{Experimental Setup}
\subsubsection{Datasets.}
To validate the effectiveness of UMAE, experiments were conducted on six datasets. The \textbf{NUS}\citep{NUS} dataset is a large-scale 3-view dataset with 30,000 samples and 31 categories, used for multi-label classification. The \textbf{YaleB}\citep{YaleB} dataset contains 3 views with 10 categories, each with 65 facial images. The \textbf{Handwritten}\citep{Handwritten} dataset consists of 6 views covering 10 categories from digits "0" to "9," with 200 samples per category. The \textbf{ROSMAP}\citep{ROSMAP} dataset includes 3 views with two categories—Alzheimer’s disease (AD) patients and normal control (NC)—with 182 and 169 samples, respectively. The \textbf{BRCA}\citep{BRCA} dataset comprises 3 views for Breast Invasive Carcinoma (BRCA) subtype classification, containing 5 categories with 46 to 436 samples per category. Lastly, the \textbf{Scene15}\citep{Scene} dataset has 3 views with 15 categories for scene classification, with each category containing 210 to 410 samples.

\subsubsection{Comparison Methods.}
This paper compares UMAE with the following methods: (1) \textbf{GCCA}\citep{GCCA}, an extension of Canonical Correlation Analysis (CCA) for handling data with more than two views; (2) \textbf{TCCA}\citep{TCCA}, which maximizes canonical correlation across multiple views to obtain a shared subspace; (3) \textbf{MVAE}\citep{MVAE}, which extends the variational autoencoder to multi-view data using a product-of-experts strategy to find a common latent subspace; (4) \textbf{MIWAE}\citep{MIWAE}, which adapts the importance-weighted autoencoder for multi-view data to impute missing data; (5) \textbf{CPM-Nets}\citep{CPM}, which directly learns joint latent representations for all views and maps these to classification predictions; (6) \textbf{DeepIMV}\citep{DeepIMV}, which applies the information bottleneck framework to extract marginal and joint representations and constructs view-specific and multi-view predictors for classification; and (7) \textbf{UIMC}\citep{UIMC}, which uses the K-nearest neighbors method to form a sampling distribution, repeatedly samples missing data, and employs the EDL fusion method to explore and leverage uncertainty from imputation for effective and reliable classification representations.

\subsection{Experimental Configuration and Construction}
We conduct extensive research on the proposed model across multiple datasets with varying missing rates, defined as \(\eta=\frac{\sum_{v=1}^V M^v}{V \times N}\). The evaluation focuses on the model's ability to handle different levels of missing data and view conflicts. To generate conflict datasets, 40\% of samples have one view randomly replaced with views from other categories. Experiments were performed on an NVIDIA RTX 4090 GPU.

\subsection{Quantitative Experimental Results}

\begin{table*}[htbp]
\centering
\vspace{-2mm}
\caption{Classification accuracy (mean$\pm$std) for different methods on various datasets under varying missing rates. \textbf{Bold} indicates the highest value, and \underline{underlined} indicates the second highest value.}

\label{tab:classification-accuracy}
\vspace{-2mm}
\small
\resizebox{\textwidth}{!}{%
\begin{tabular}{l c c c c c c c}
\toprule
\multirow{2}{*}{\textbf{Datasets}} &
\multirow{2}{*}{\textbf{Methods}} &
\multicolumn{6}{c}{\textbf{Missing Rates}} \\
\cmidrule(lr){3-8}
& &
{$\eta=0$} &
{$\eta=0.1$} &
{$\eta=0.2$} &
{$\eta=0.3$} &
{$\eta=0.4$} &
{$\eta=0.5$} \\
\midrule
\multirow{8}{*}{YaleB} &
GCCA & 0.9692$\pm$0.00 & 0.9385$\pm$0.01 & 0.9077$\pm$0.02 & 0.8615$\pm$0.03 & 0.8385$\pm$0.02 & 0.8231$\pm$0.02 \\
& TCCA & 0.9846$\pm$0.00 & 0.9625$\pm$0.00 & 0.9492$\pm$0.01 & 0.9077$\pm$0.01 & 0.8846$\pm$0.02 & 0.8615$\pm$0.01 \\
& MVAE & \textbf{1.0000$\pm$0.00} & \underline{0.9969$\pm$0.00} & 0.9861$\pm$0.00 & 0.9831$\pm$0.01 & 0.9692$\pm$0.02 & 0.9599$\pm$0.01 \\
& MIVAE & \textbf{1.0000$\pm$0.00}  & 0.9923$\pm$0.00 & 0.9923$\pm$0.01 & 0.9903$\pm$0.01 & 0.9846$\pm$0.01 & 0.9692$\pm$0.03 \\
& CPM-Nets & \underline{0.9915$\pm$0.02} & 0.9862$\pm$0.01 & 0.9800$\pm$0.01 & 0.9700$\pm$0.02 & 0.9469$\pm$0.01 & 0.9100$\pm$0.02 \\
& DeepIMV & \textbf{1.0000$\pm$0.00}  & 0.9846$\pm$0.03 & 0.9231$\pm$0.02 & 0.9154$\pm$0.08 & 0.8923$\pm$0.02 & 0.8718$\pm$0.06 \\
& UIMC & \textbf{1.0000$\pm$0.00}  & \textbf{1.0000$\pm$0.00}  & \textbf{0.9981$\pm$0.00} & \underline{0.9962$\pm$0.01} & \underline{0.9847$\pm$0.01} & \underline{0.9769$\pm$0.01} \\
\cmidrule{2-8}
& Ours & \textbf{1.0000$\pm$0.00} & \textbf{1.0000$\pm$0.00} & \underline{0.9969$\pm$0.00} & \textbf{0.9965$\pm$0.00} & \textbf{0.9848$\pm$0.01} & \textbf{0.9831$\pm$0.01} \\
\midrule
\multirow{8}{*}{ROSMAP} &
GCCA & 0.6953$\pm$0.03 & 0.6571$\pm$0.02 & 0.6429$\pm$0.04 & 0.6143$\pm$0.03 & 0.5714$\pm$0.06 & 0.5429$\pm$0.06 \\
& TCCA & 0.7143$\pm$0.03 & 0.7072$\pm$0.01 & 0.6857$\pm$0.05 & 0.6500$\pm$0.05 & 0.6286$\pm$0.03 & 0.6036$\pm$0.06 \\
& MVAE & 0.7429$\pm$0.02 & 0.7286$\pm$0.05 & 0.7143$\pm$0.05 & 0.6786$\pm$0.03 & 0.6786$\pm$0.06 & 0.6524$\pm$0.06 \\
& MIVAE & 0.7429$\pm$0.03 & 0.7286$\pm$0.02 & 0.6714$\pm$0.05 & 0.6571$\pm$0.03 & 0.6571$\pm$0.05 & 0.6357$\pm$0.08 \\
& CPM-Nets & 0.7840$\pm$0.05 & 0.7517$\pm$0.04 & 0.7394$\pm$0.06 & 0.7183$\pm$0.04 & 0.6901$\pm$0.08 & 0.6409$\pm$0.08 \\
& DeepIMV & 0.7607$\pm$0.03 & 0.7429$\pm$0.01 & 0.7143$\pm$0.05 & 0.6643$\pm$0.05 & 0.6524$\pm$0.06 & 0.6250$\pm$0.06 \\
& UIMC & \underline{0.8714$\pm$0.00} & \underline{0.8429$\pm$0.03} &\underline{ 0.7714$\pm$0.05} & \underline{0.7464$\pm$0.03} &\underline{ 0.7214$\pm$0.03} & \underline{0.7143$\pm$0.02} \\
\cmidrule{2-8}
& Ours & \textbf{0.8776$\pm$0.02} & \textbf{0.8501$\pm$0.02} & \textbf{0.7936$\pm$0.02} & \textbf{0.7795$\pm$0.03} & \textbf{0.7377$\pm$0.04} & \textbf{0.7297$\pm$0.02} \\
\midrule
\multirow{8}{*}{Handwritten} &
GCCA & 0.9500$\pm$0.01 & 0.9350$\pm$0.02 & 0.9100$\pm$0.01 & 0.8875$\pm$0.02 & 0.8425$\pm$0.02 & 0.8200$\pm$0.03 \\
& TCCA & 0.9725$\pm$0.00 & 0.9650$\pm$0.00 & 0.9575$\pm$0.02 & 0.9350$\pm$0.01 & 0.9200$\pm$0.01 & 0.9100$\pm$0.02 \\
& MVAE & 0.9800$\pm$0.00 & 0.9750$\pm$0.01 & 0.9700$\pm$0.00 & 0.9650$\pm$0.01 & 0.9575$\pm$0.01 & 0.9500$\pm$0.01 \\
& MIVAE & 0.9800$\pm$0.00 & 0.9800$\pm$0.00 & 0.9725$\pm$0.00 & 0.9650$\pm$0.00 & 0.9475$\pm$0.01 & 0.9375$\pm$0.02 \\
& CPM-Nets & 0.9550$\pm$0.01 & 0.9475$\pm$0.01 & 0.9375$\pm$0.01 & 0.9300$\pm$0.02 & 0.9225$\pm$0.01 & 0.9125$\pm$0.01 \\
& DeepIMV & \underline{0.9908$\pm$0.04} & \underline{0.9883$\pm$0.02} & \textbf{0.9850$\pm$0.04}  & 0.9750$\pm$0.02 & 0.9625$\pm$0.04 & 0.9450$\pm$0.06 \\
& UIMC & 0.9825$\pm$0.00 & 0.9800$\pm$0.00 & 0.9800$\pm$0.00 & \underline{0.9775$\pm$0.00} & \underline{0.9700$\pm$0.01} & \underline{0.9600$\pm$0.01} \\
\cmidrule{2-8}
& Ours & \textbf{0.9915$\pm$0.00} & \textbf{0.9913$\pm$0.00} & \underline{0.9826$\pm$0.00} & \textbf{0.9776$\pm$0.00} & \textbf{0.9705$\pm$0.00} & \textbf{0.9605$\pm$0.00} \\
\midrule
\multirow{8}{*}{BRCA} &
GCCA & 0.7371$\pm$0.03 & 0.7143$\pm$0.03 & 0.6971$\pm$0.04 & 0.6762$\pm$0.02 & 0.6514$\pm$0.03 & 0.6381$\pm$0.04 \\
& TCCA & 0.7543$\pm$0.02 & 0.7314$\pm$0.03 & 0.7238$\pm$0.04 & 0.7129$\pm$0.03 & 0.6857$\pm$0.04 & 0.6743$\pm$0.03 \\
& MVAE & 0.7885$\pm$0.03 & 0.7691$\pm$0.02 & 0.7347$\pm$0.01 & 0.6968$\pm$0.03 & 0.6633$\pm$0.05 & 0.6388$\pm$0.03 \\
& MIVAE & 0.7885$\pm$0.02 & 0.7352$\pm$0.03 & 0.7314$\pm$0.03 & 0.7105$\pm$0.02 & 0.7029$\pm$0.02 & \underline{0.6857$\pm$0.04} \\
& CPM-Nets & 0.7388$\pm$0.02 & 0.7317$\pm$0.04 & 0.7107$\pm$0.08 & 0.7233$\pm$0.04 & 0.6980$\pm$0.05 & 0.6788$\pm$0.03 \\
& DeepIMV & 0.7686$\pm$0.03 & 0.7614$\pm$0.02 & 0.7457$\pm$0.02 & 0.7414$\pm$0.02 & 0.7400$\pm$0.02 & 0.6714$\pm$0.04 \\
& UIMC & \underline{0.8286$\pm$0.01} & \underline{0.7943$\pm$0.01} & \underline{0.7771$\pm$0.01} & \underline{0.7657$\pm$0.02} & \underline{0.7543$\pm$0.02} & \textbf{0.7429$\pm$0.02} \\
\cmidrule{2-8}
& Ours & \textbf{0.8289$\pm$0.01} & \textbf{0.8008$\pm$0.02} & \textbf{0.7942$\pm$0.01} & \textbf{0.7772$\pm$0.02} & \textbf{0.7607$\pm$0.02} & \textbf{0.7429$\pm$0.02} \\
\midrule
\multirow{8}{*}{Scene15} &
GCCA & 0.6611$\pm$0.02 & 0.6511$\pm$0.01 & 0.6176$\pm$0.01 & 0.5708$\pm$0.01 & 0.5385$\pm$0.02 & 0.5006$\pm$0.02 \\
& TCCA & 0.6878$\pm$0.02 & 0.6644$\pm$0.01 & 0.6566$\pm$0.01 & 0.6187$\pm$0.01 & 0.5741$\pm$0.01 & 0.5563$\pm$0.02 \\
& MVAE & 0.7681$\pm$0.00 & 0.7346$\pm$0.01 & 0.7157$\pm$0.01 & 0.6689$\pm$0.01 & 0.6444$\pm$0.01 & 0.6098$\pm$0.01 \\
& MIVAE & 0.7681$\pm$0.03 & 0.7179$\pm$0.01 & 0.6990$\pm$0.01 & 0.6566$\pm$0.01 & 0.6265$\pm$0.02 & 0.5875$\pm$0.02 \\
& CPM-Nets & 0.6990$\pm$0.02 & 0.6566$\pm$0.02 & 0.6388$\pm$0.00 & 0.6265$\pm$0.01 & 0.5903$\pm$0.01 & 0.5708$\pm$0.01 \\
& DeepIMV & 0.7124$\pm$0.00 & 0.6934$\pm$0.02 & 0.6656$\pm$0.01 & 0.6410$\pm$0.00 & 0.5853$\pm$0.02 & 0.5719$\pm$0.01 \\
& UIMC & \underline{0.7770$\pm$0.00} &\underline{ 0.7581$\pm$0.01} & \underline{0.7347$\pm$0.00} & \underline{0.6990$\pm$0.01} & \underline{0.6689$\pm$0.01} & \underline{0.6254$\pm$0.02} \\
\cmidrule{2-8}
& Ours & \textbf{0.7712$\pm$0.01} & \textbf{0.7603$\pm$0.01} & \textbf{0.7401$\pm$0.01} & \textbf{0.7018$\pm$0.01} & \textbf{0.6800$\pm$0.00} & \textbf{0.6453$\pm$0.01} \\
\bottomrule
\vspace{-2mm}
\end{tabular}%
}
\end{table*}

\begin{table*}[htbp]
\centering
\vspace{-2mm}
\caption{Classification accuracy (mean$\pm$std) for different methods on conflict datasets under varying missing rates.}
\label{tab:ZIMP,MIMP,UMAE..}
\resizebox{\textwidth}{!}{ 
\begin{tabular}{llcccccc}
\toprule
\textbf{Dataset} & \textbf{Method} & \multicolumn{6}{c}{\textbf{Missing Rates ($\eta$)}} \\
\cmidrule(lr){3-8}
& & \textbf{$\eta=0$} & \textbf{$\eta=0.1$} & \textbf{$\eta=0.2$} & \textbf{$\eta=0.3$} & \textbf{$\eta=0.4$} & \textbf{$\eta=0.5$} \\
\midrule
\multirow{5}{*}{Handwritten} 
& ZIMP & 0.9725$\pm$0.00 & 0.9691$\pm$0.00 & 0.9613$\pm$0.01 & 0.9484$\pm$0.01 & 0.9212$\pm$0.01 & 0.9143$\pm$0.02 \\
& MIMP & 0.9805$\pm$0.00 & 0.9776$\pm$0.00 & 0.9669$\pm$0.01 & 0.9453$\pm$0.01 & 0.9234$\pm$0.01 & 0.9158$\pm$0.01 \\
& UMAE-F & 0.9813$\pm$0.00 & 0.9800$\pm$0.00 & 0.9723$\pm$0.00 & 0.9468$\pm$0.01 & 0.9256$\pm$0.01 & 0.9208$\pm$0.01 \\
& UMAE-V & 0.9915$\pm$0.00 & 0.9912$\pm$0.00 & 0.9779$\pm$0.00 & 0.9750$\pm$0.00 & 0.9659$\pm$0.00 & 0.9536$\pm$0.00 \\
& \textbf{UMAE-J} & \textbf{0.9915$\pm$0.00} & \textbf{0.9913$\pm$0.00} & \textbf{0.9826$\pm$0.00} & \textbf{0.9776$\pm$0.00} & \textbf{0.9705$\pm$0.00} & \textbf{0.9605$\pm$0.00} \\
\midrule
\multirow{5}{*}{ROSMAP} 
& ZIMP & 0.7814$\pm$0.02 & 0.7684$\pm$0.01 & 0.7299$\pm$0.03 & 0.7154$\pm$0.03 & 0.6991$\pm$0.04 & 0.6728$\pm$0.05 \\
& MIMP & 0.8167$\pm$0.02 & 0.8204$\pm$0.01 & 0.7543$\pm$0.02 & 0.7243$\pm$0.02 & 0.7164$\pm$0.02 & 0.6934$\pm$0.03 \\
& UMAE-F & 0.8218$\pm$0.03 & 0.8173$\pm$0.02 & 0.7601$\pm$0.02 & 0.7413$\pm$0.02 & 0.7156$\pm$0.02 & 0.7158$\pm$0.05 \\
& UMAE-V & 0.8767$\pm$0.02 & 0.8427$\pm$0.02 & 0.7829$\pm$0.03 & 0.7708$\pm$0.02 & 0.7206$\pm$0.04 & 0.7285$\pm$0.03 \\
& \textbf{UMAE-J} & \textbf{0.8776$\pm$0.02} & \textbf{0.8501$\pm$0.02} & \textbf{0.7936$\pm$0.02} & \textbf{0.7795$\pm$0.03} & \textbf{0.7377$\pm$0.04} & \textbf{0.7297$\pm$0.02} \\
\bottomrule

\end{tabular}
}
\vspace{-4mm}
\end{table*}
\subsubsection{Comparison on Normal Datasets.}  
We first evaluated model effectiveness using standard datasets with varying missing rates $\eta = [0, 0.1, 0.2, 0.3, 0.4, 0.5]$. All methods were implemented using the same network architecture. As presented in Table\ref{tab:classification-accuracy}, UMAE consistently outperformed all existing methods, including the state-of-the-art UIMC. This is particularly notable as the K-nearest neighbors imputation method used in UIMC showed diminishing effectiveness with sparse samples, larger sample distances, and higher missing rates. UMAE’s superior performance can be attributed to its learning-based imputation approach, which operates on feature layers of missing views and robustly leverages global information to generate high-quality imputed views, thereby achieving leading-edge performance.

\subsubsection{Comparison on Conflictive Datasets.}
\begin{figure}[hbtp]
    \centering
    \includegraphics[width=0.5\textwidth]{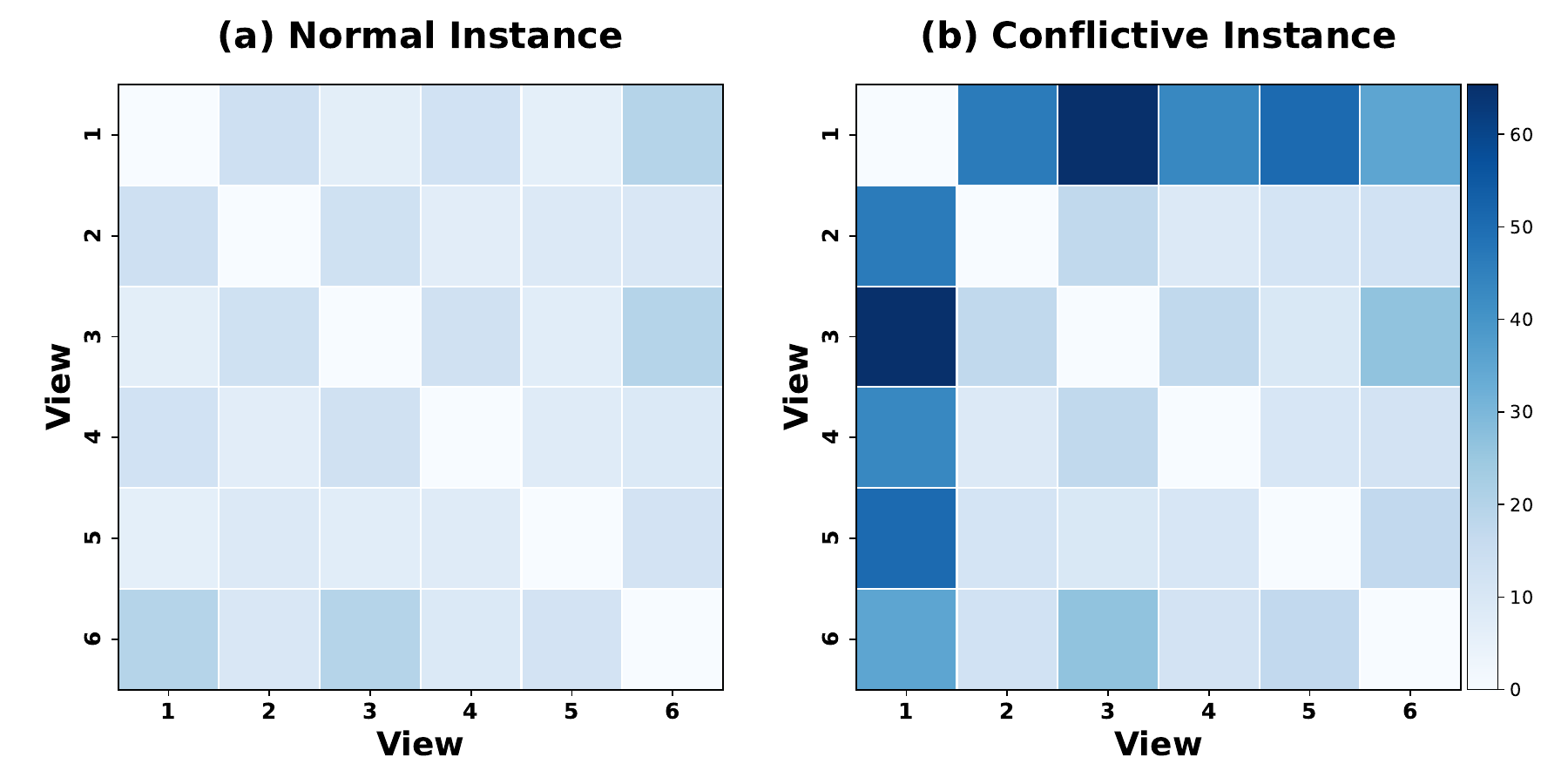}
    \caption{Conflictive degree visualization.}
    \label{fig:heatmap}
\end{figure}

\begin{figure}[hbtp]
    \centering
    \includegraphics[width=\columnwidth]{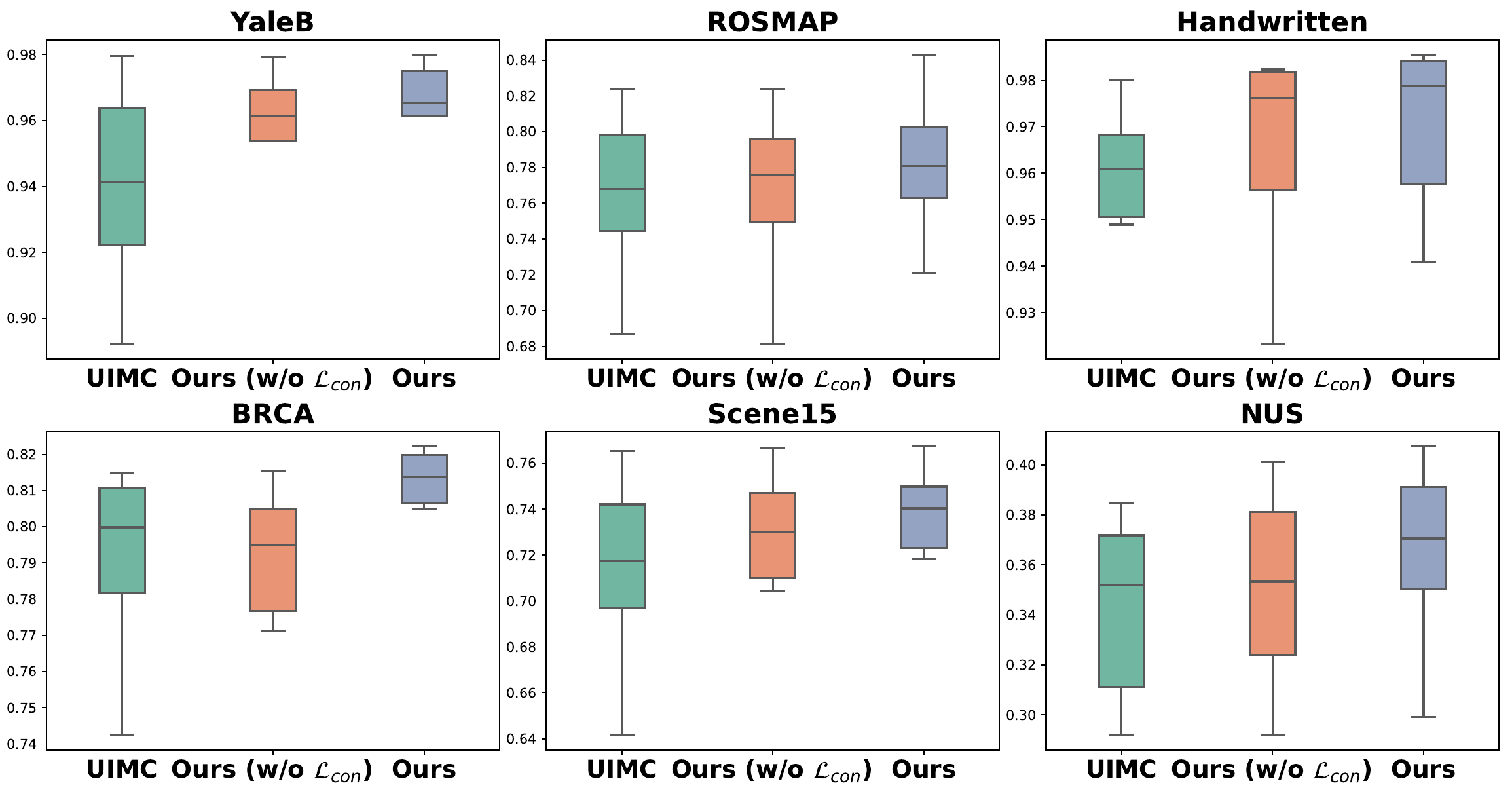}
    \vspace{-4mm}
    \caption{Robustness to conflict: Performance comparison on conflict datasets with varying missing rates across different models.}
    \label{fig:xiangxingtu}
    \vspace{-5mm}
\end{figure}

To validate the effectiveness of the proposed conflict loss module,we conducted experiments on conflict datasets.Fig. \ref{fig:heatmap} visually compares the degrees of conflict across six views on the Handwritten dataset in both normal and conflict scenarios that are only replaced in the first view.These visual comparisons demonstrate the conflict measurement capability of the loss module.We evaluated UIMC, our method without the $\mathcal{L}_{\text{con}}$, and our method with the $\mathcal{L}_{\text{con}}$ on conflict datasets with various missing rates, and we plotted the results as box plots.As shown in Fig. \ref{fig:xiangxingtu} , our conflict measurement method effectively captures the conflicts, and the conflict loss module significantly enhances the model's robustness, effectively reducing conflict risks in the DSCR.

\subsubsection{Ablation Study.}  
To evaluate each stage of APLN, we conducted experiments using Zero-Imputation (ZIMP) and Mean-Imputation (MIMP) as baselines, alongside UMAE-F, UMAE-V, and UMAE-J.As shown in Table \ref{tab:ZIMP,MIMP,UMAE..}, UMAE-F with its noise imputation method at the feature layer, achieve competitive performance with regard to baseline methods. This stage primarily enables a coarse feature alignment function, allowing UMAE-V to refine the aligned feature domain. In UMAE-V, vae and edl module is learned, significantly enhancing the model at this stage. UMAE-J, where the feature layers $\{f_c^v\}_{v=1}^V$,VAE and EDL moudle are jointly trained, improves coordination and adaptability among model components, leading to overall performance optimization.

\subsection{Qualitative Experimental Results}
\begin{figure}[hbtp]
    \centering
    \includegraphics[width=\columnwidth]{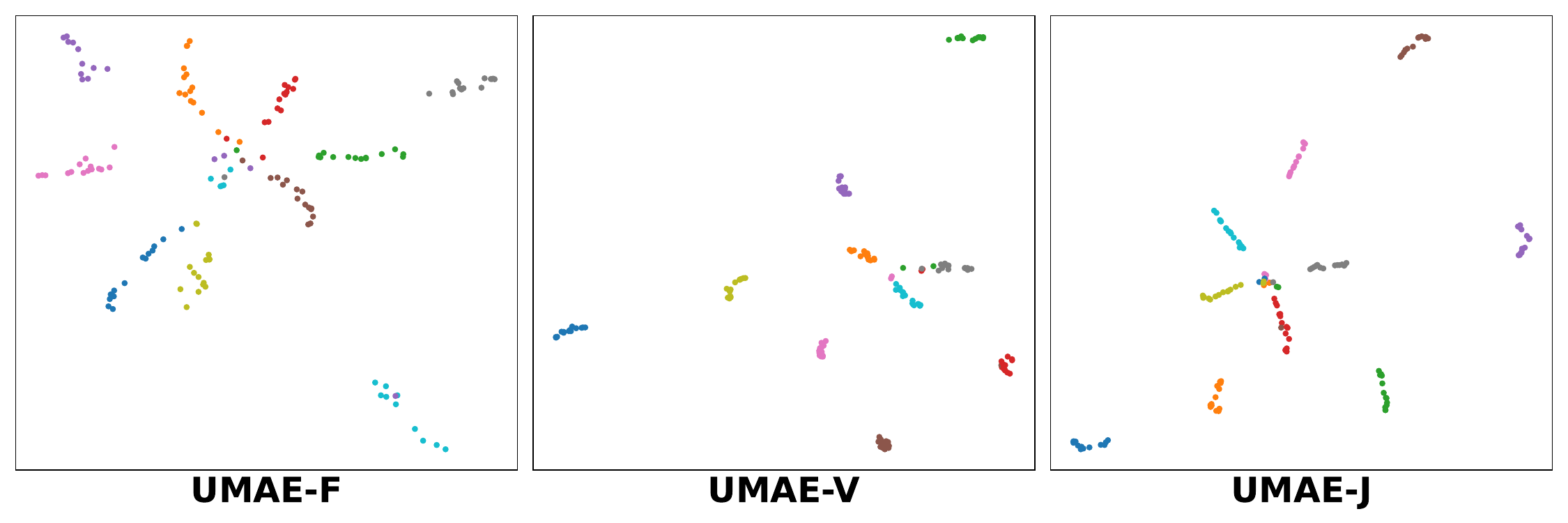}
    \vspace{-2mm}
    \caption{T-SNE visualization of Evidence distributions across different stages of the APLN on the Handwritten dataset. UMAE-F represents the feature training stage with incomplete views, UMAE-V shows the distribution after the view-specific alignment, and UMAE-J illustrates the joint evidence distribution after final integration.}
    \label{fig:tsne}
    \vspace{-4mm}
\end{figure}

\begin{figure}[hbtp]
    \centering
    \begin{minipage}{0.49\linewidth} 
        \centering
        \includegraphics[width=\textwidth]{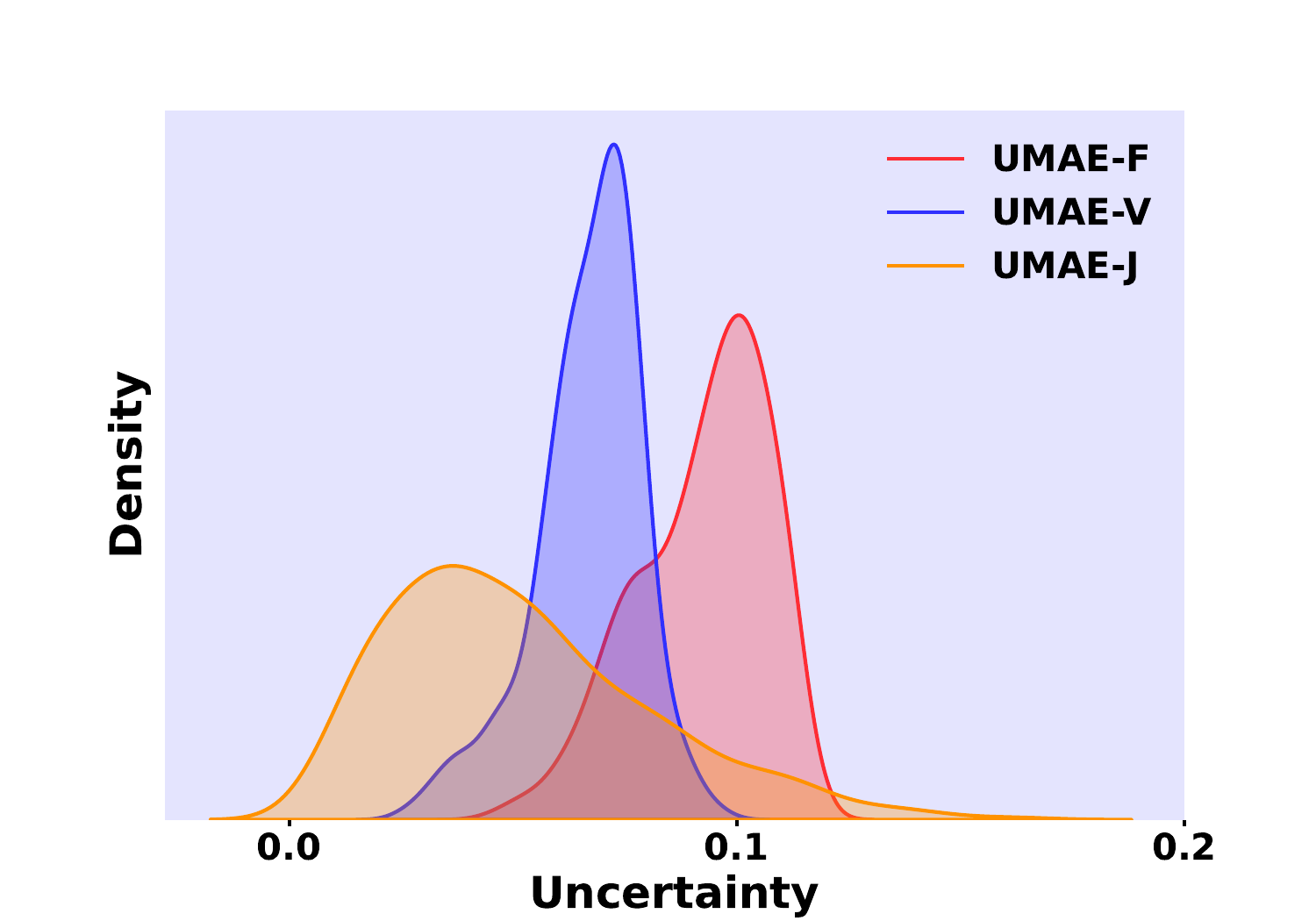}
        \caption*{(a) YaleB}
    \end{minipage}%
    \hspace{0.01\linewidth} 
    \begin{minipage}{0.49\linewidth}
        \centering
        \includegraphics[width=\textwidth]{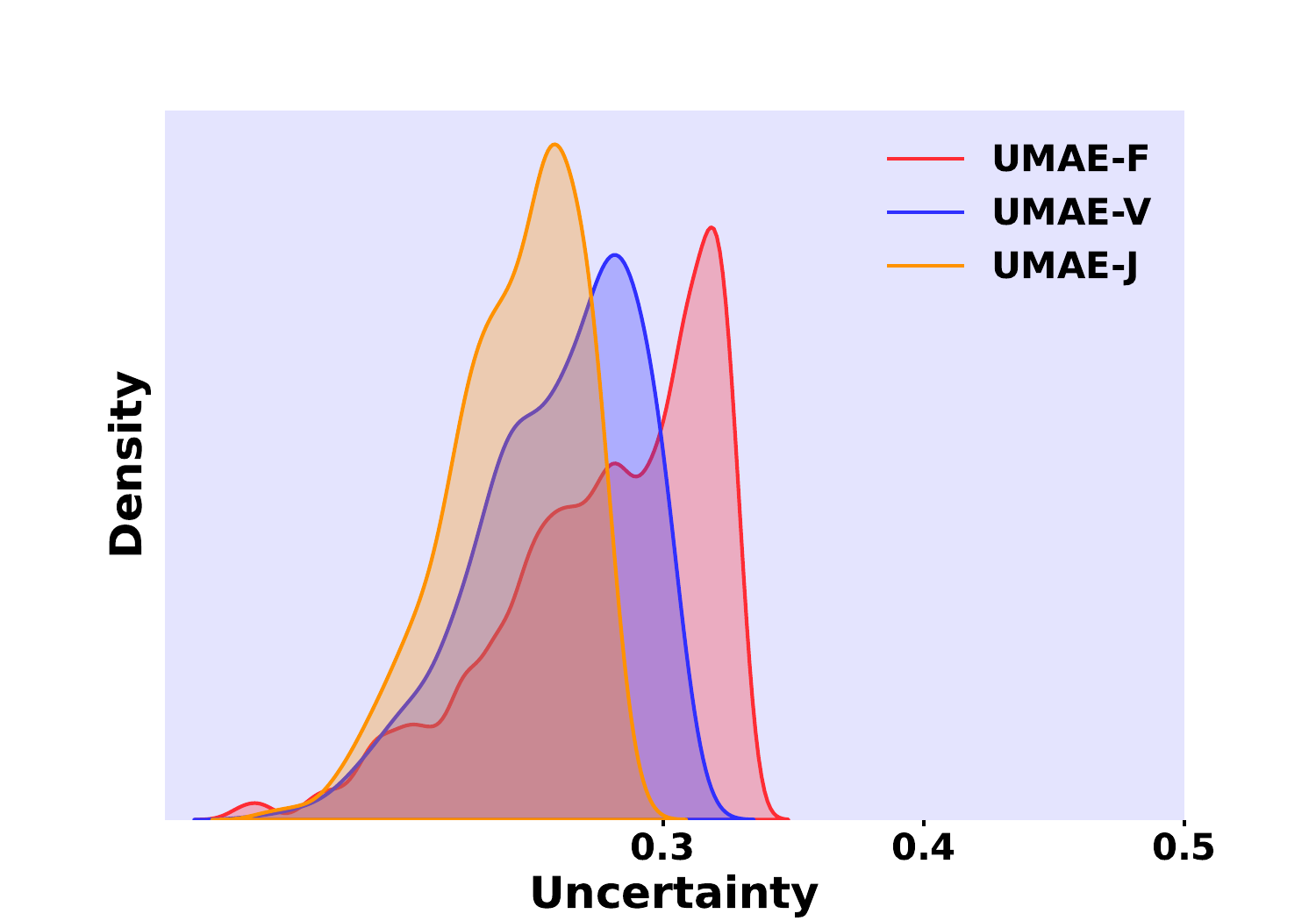}
        \caption*{(b) Scene15}
    \end{minipage}%
    \vspace{0.01\linewidth} 
    \begin{minipage}{0.49\linewidth}
        \centering
        \includegraphics[width=\textwidth]{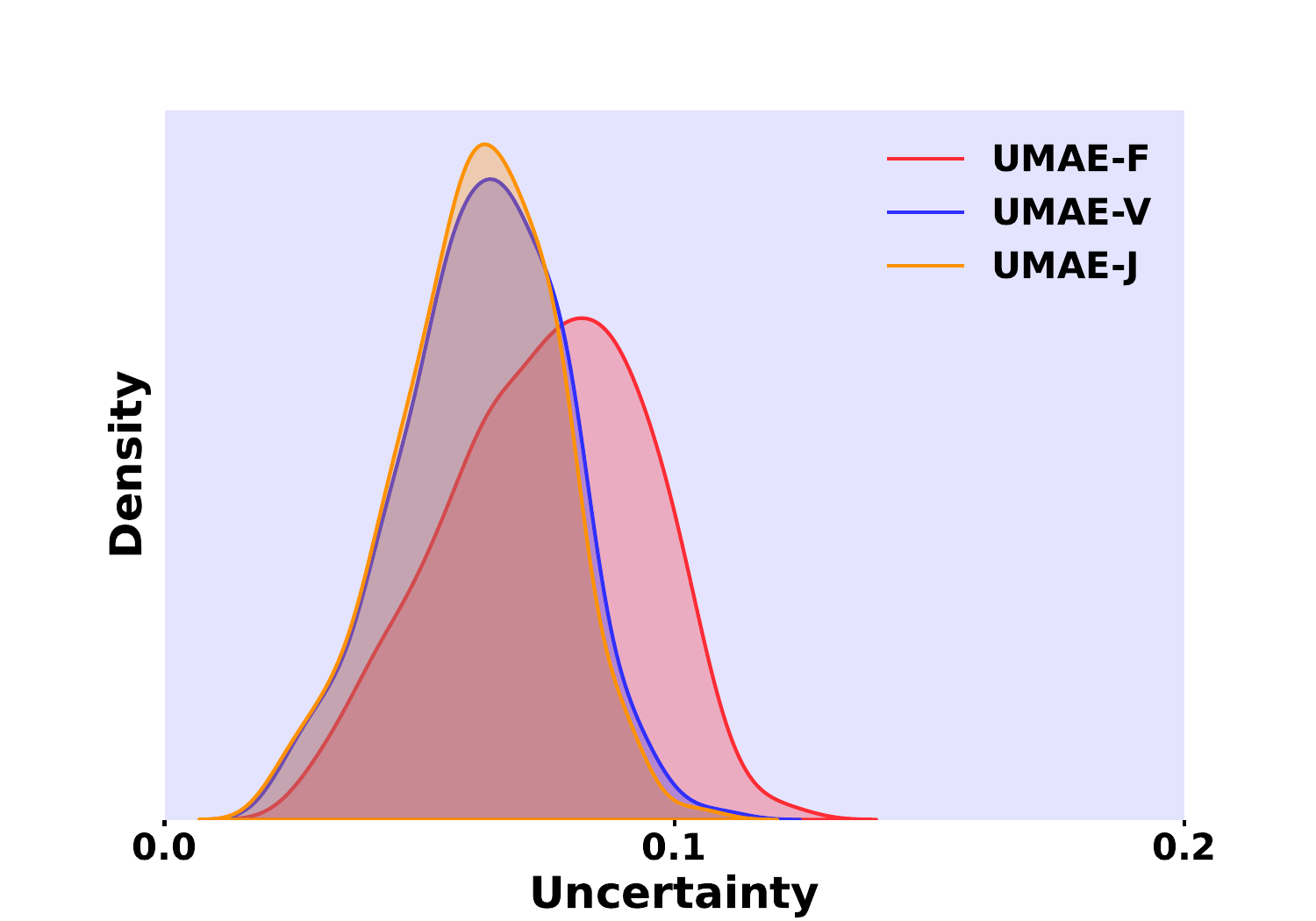}
        \caption*{(c) Handwritten}
    \end{minipage}%
    \hspace{0.01\linewidth} 
    \begin{minipage}{0.49\linewidth}
        \centering
        \includegraphics[width=\textwidth]{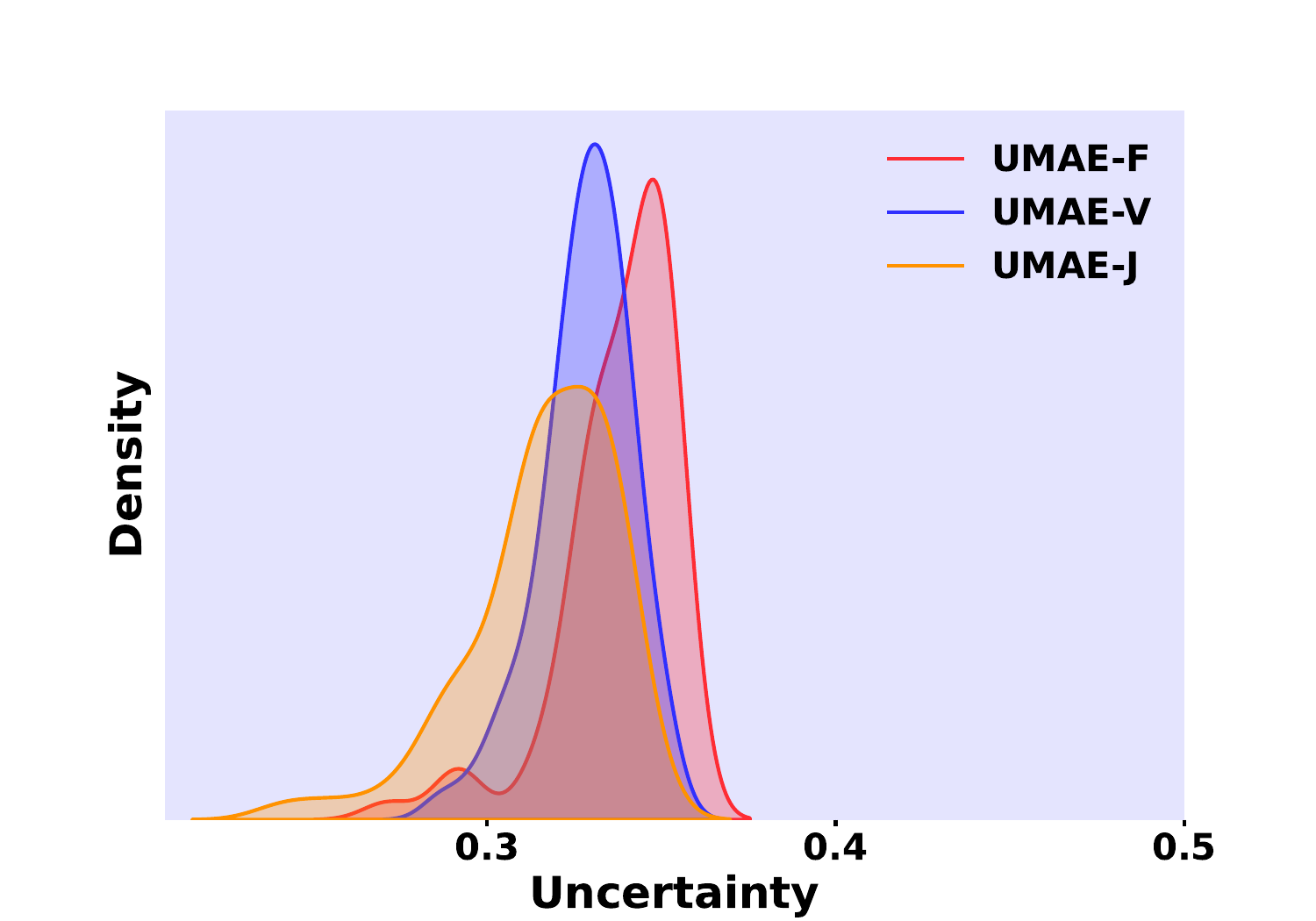}
        \caption*{(d) ROSMAP}
    \end{minipage}
    \caption{KDE plots showing the uncertainty distribution across the three stages of APLN training for four datasets: YaleB, Scene15, Handwritten, and ROSMAP. UMAE-F represents the feature traning stage, UMAE-V shows the view-specific alignment stage, and UMAE-J illustrates the final joint evidence integration stage.}
    \label{fig:kde}
    \vspace{-2mm}
\end{figure}

\subsubsection{t-SNE Visualization of Model Stages.}  
To visually and qualitatively assess the effectiveness of our approach across the three stages, we conducted a t-SNE experiment on the Handwritten dataset. As illustrated in Fig. \ref{fig:tsne}, the t-SNE visualizations demonstrate the progressive improvement of the model's classification abilities across the stages of APLN, ultimately achieving state-of-the-art results.

\subsubsection{Kernel Density Estimation  Analysis of Uncertainty Across Model Stages.} 
To quantitatively and qualitatively assess the reduction in output uncertainty across the three stages of our approach, we performed a kernel density estimation (KDE) analysis on the 4 datasets under a missing rate of $\eta = 0.4$. As shown in Fig. \ref{fig:kde}, the KDE plots clearly illustrate the progressive decrease in model output uncertainty as the APLN process unfolds. These findings underscore the effectiveness of our method in refining the model's confidence in its predictions over the course of the three stages.

\section{Conclusion}
In this work, we introduced the APLN to address challenges in incomplete multi-view classification, focusing on mitigating bias and managing uncertainty in corrupted observed domains. Our approach refines imputation in the latent space and integrates a conflict-aware DSCR to enhance decision-making robustness. Experimental results on benchmark datasets demonstrate that APLN significantly outperforms traditional methods, particularly in environments with high uncertainty and conflicting evidence. However, the computational demands of our method, especially during the progressive learning and evidence fusion stages, suggest potential challenges in scaling to very large datasets. Future work will focus on improving the computational efficiency of APLN, making it more accessible for larger-scale applications while maintaining its robustness in IMVC.


\bibliography{aaai25}
\end{document}